\documentclass[11pt,a4paper]{article}
\usepackage[hyperref]{acl2020}
\usepackage{times}
\usepackage{latexsym}

\usepackage{microtype}
\usepackage{algorithm}
\usepackage{algorithmic}
\usepackage{amsmath}
\usepackage{amsthm}
\usepackage{bbm}
\usepackage{graphicx}

\aclfinalcopy 

\title{Self-Distillation Mixup Training for Non-autoregressive Neural Machine Translation}

\author{Jiaxin Guo\textsuperscript{\rm 1}, 
  Minghan Wang\textsuperscript{\rm 1}, 
  Daimeng Wei\textsuperscript{\rm 1},
  Hengchao Shang\textsuperscript{\rm 1},
  Yuxia Wang\textsuperscript{\rm 2},
  \\
  {\bf Zongyao Li\textsuperscript{\rm 1}},
  {\bf Zhengzhe Yu\textsuperscript{\rm 1}},
  {\bf Zhanglin Wu\textsuperscript{\rm 1}},
  {\bf Yimeng Chen\textsuperscript{\rm 1},}
  {\bf Chang Su\textsuperscript{\rm 1},}
  \\
  {\bf Min Zhang\textsuperscript{\rm 1},}
  {\bf Lizhi Lei\textsuperscript{\rm 1},}
  {\bf Shimin Tao\textsuperscript{\rm 1},}
  {\bf Hao Yang\textsuperscript{\rm 1}}
  \\
  \textsuperscript{\rm 1}Huawei Translation Services Center, Beijing, China \\
  \textsuperscript{\rm 2}The University of Melbourne, Melbourne, Australia\\
  \tt \{guojiaxin1,wangminghan,weidaimeng,shanghengchao,\\ 
  \tt lizongyao,yuzhengzhe,wuzhanglin2,chenyimeng,suchang8,\\
  \tt zhangmin186,leilizhi,taoshimin,yanghao30\}@huawei.com \\
  \texttt{yuxiaw@student.unimelb.edu.au}}

\date{}

\begin{document}
\maketitle
\begin{abstract}
Recently, non-autoregressive (NAT) models predict outputs in parallel, achieving substantial improvements in generation speed compared to autoregressive (AT) models.  While performing worse on raw data, most NAT models are trained as student models on distilled data generated by AT teacher models, which is known as sequence-level Knowledge Distillation. An effective training strategy to improve the performance of AT models is Self-Distillation Mixup (SDM) Training, which pre-trains a model on raw data, generates distilled data by the pre-trained model itself and finally re-trains a model on the combination of raw data and distilled data. In this work, we aim to view SDM for NAT models, but find directly adopting SDM to NAT models gains no improvements in terms of translation quality. Through careful analysis, we observe the invalidation is correlated to $\textit{Modeling Diversity}$ and $\textit{Confirmation Bias}$ between the AT teacher model and the NAT student models. Based on these findings, we propose an enhanced strategy named SDMRT by adding two stages to classic SDM: one is Pre-Rerank on self-distilled data, the other is Fine-Tune on Filtered teacher-distilled data. Our results outperform baselines by 0.6$\sim$1.2 BLEU on multiple NAT models. As another bonus, for Iterative Refinement NAT models, our methods can outperform baselines within half iteration number, which means 2$\times$ acceleration.
\end{abstract}

\section{Introduction}

Autoregressive (AT) neural machine translation \citep{DBLP:journals/corr/BahdanauCB14} models decode tokens one-by-one, which ensure the robustness of the intrinsic language model but make the inference slow ({\S}\ref{sec:AT}). Recently, non-autoregressive (NAT) models generate all outputs in parallel.
It speeds up the inference, but at the cost of breaking the dependency between adjacent tokens, leading to worse performance than standard AT models ({\S}\ref{sec:NAT}). 

Knowledge distillation \citep{DBLP:journals/corr/HintonVD15} is a method that trains a student model to perform better by learning from a stronger teacher model. This method has been proved to be necessary for NAT models training in \citep{DBLP:conf/iclr/ZhouGN20}, and most NAT models are trained on distilled data generated by an AT teacher model to achieve competitive performance. This method of distilled data generation is called sequence-level knowledge distillation \citep{DBLP:conf/emnlp/KimR16} ({\S}\ref{sec:KD}).

When the teacher model is the same as the student model, we call it Self-Distillation. Using the combination of the original data and the self distilled data to re-train the model itself is Self-Distillation Mixup (SDM) Training. \citep{DBLP:journals/corr/FreitagAS17} has proved that SDM can improve the performance of AT models ({\S}\ref{sec:SDM}).

In this work, we study a general training strategy to improve the NAT performance. A promising attempt to address this issue is to improve the AT teacher model by using some strategies such as SDM Training. The problem is that, higher quality distilled data generated by a better AT teacher model may not improve the performance of NAT student models generally ({\S}\ref{sec:prob_1}). Another approach is to adopt SDM Training to NAT models directly. But the second problem occurs that, classic SDM Training has no effect when NAT models are trained on AT-distilled data ({\S}\ref{sec:prob_2}).

We propose an enhanced training progress named \textit{Self-Distillation Mixup Training with Pre-Rerank and Fine-Tune} (SDMRT). In the stage of distilled data generation, we use Rerank Algorithm \citep{DBLP:conf/acl/0001AR20} to reduce the \textit{Modeling Diversity} between AT and NAT models. In the stage of model training, we Fine-Tune the NAT models on the filtered data to reduce the \textit{Confirmation Bias} ({\S}\ref{sec:approach}). 

Our contributions are as follows:

\begin{itemize}
    \item We first apply SDM to NAT training and introduce an enhanced training strategy named SDMRT to significantly improve translation quality of multiple NAT models ({\S}\ref{sec:method}; {\S}\ref{sec:main_results}).
    \item We analyse why classic SDM fails on NAT models trained on AT-distilled data. Specifically, we theoretically demonstrate $\textit{Modeling Diversity}$ and $\textit{Confirmation Bias}$ between the AT teacher model and the NAT student models ({\S}\ref{sec:analysis}; {\S}\ref{sec:revisit-modeling}; {\S}\ref{sec:revisit-bias}).
    \item We carefully study the effect of different Rerank strategies on self-distilled data and Data-filtering methods on teacher-distilled data. Furthermore, we find our SDMRT can be regarded as an acceleration method to reduce iterations for Iterative Refinement NAT models ({\S}\ref{sec:speed_irnat}; {\S}\ref{sec:analysis_rerank}; {\S}\ref{sec:analysis_filter}).
\end{itemize}

\section{Background}
\subsection{Autoregressive NMT Models}
\label{sec:AT}
Let $x$ denote the input sequence and $y$ denote the output sequence. Let $\mathcal{D}$ denote the original parallel data: $\mathcal{D} = \{\mathcal{X}; \mathcal{Y}\}$. In order to model the joint probability of the output sequence, autoregressive (AT) models usually generate each output token $y_t$ conditioned on the previously generated ones\citep{DBLP:journals/corr/BahdanauCB14}. The formulation is as following:
\begin{align}
    p(y|x) = \prod_{t=1}^T p(y_t|\textbf{y}_{<t},x)
\end{align}
\subsection{Non-autoregressive NMT Model}
\label{sec:NAT}
Non-autoregressive (NAT) models \citep{DBLP:conf/icml/DuTJ21,DBLP:conf/acl/GuK21,DBLP:conf/emnlp/SahariaCSN20} generate all outputs in parallel and break the dependency between the output tokens. The basic formulation of a NAT model independently factors the conditional distribution:

\begin{align}
    p(y|x) = \prod_{t=1}^T p(y_t|x)
\end{align}

According to the decoding strategy, NAT can be classified into two types, fully non-autoregressive models and iterative refinement models:

\paragraph{Fully Non-autoregressive models}
\begin{itemize}
    \item \textbf{Vanilla NAT} \citep{DBLP:conf/iclr/Gu0XLS18} Vanilla NAT (V-NAT) is the first NAT model. We use the simplified version where the decoder’s inputs are directly copied from the encoder without considering latent variables.
    \item \textbf{NAT-CRF} \citep{DBLP:conf/nips/SunLWHLD19} NAT-CRF designs an efficient approximation for Conditional Random Fields (CRF) for decoder's outputs.
\end{itemize}

\paragraph{Iterative Refinement (IR) NAT models} 

\begin{itemize}
    \item \textbf{iNAT} \citep{DBLP:conf/emnlp/LeeMC18} iNAT extends the vanilla NAT by iteratively refining the translation. At each refinement step, the outputs of the previous step would be taken as inputs.
    \item \textbf{CMLM} \citep{DBLP:conf/emnlp/GhazvininejadLL19} CMLM adopts a masked language model to progressively generate the sequence from entirely masked inputs. At each refinement step, the outputs of the previous step would be partially re-masked and taken as inputs.
\end{itemize}

\subsection{Knowledge Distillation}
\label{sec:KD}
Knowledge distillation \citep{DBLP:journals/corr/HintonVD15} was originally proposed for training a weaker student classifier on the targets predicted from a stronger teacher model. A typical approach is using the label probabilities produced by the teacher as “soft targets” $q_i = \exp (z_i /\tau) \big/ \sum_j\exp (z_j/\tau)$ for training the student model, where $q_i$ and $z_i$ are the probability and the logit of class $i$ respectively and $\tau$ is the temperature. Prior work has shown the effectiveness of adopting knowledge distillation in adversarial defense, neural network compression, and fast inference for speech synthesis.

\paragraph{Seq-Level KD} In the context of sequence generation, \citet{DBLP:conf/emnlp/KimR16} extend knowledge distillation to the sentence level using “hard targets” from a pretrained large teacher model to train a small sequence generation model. More precisely, the teacher distribution $q(t|x)$ is approximated by its mode: $q(t|x) \approx \mathbbm{1} \{ t = \mathop{\arg\max}\limits_{t \in \mathcal{T}} q(t|x)\}$ with the following objectives:

\begin{equation}
\begin{aligned}
    \mathcal{L}_{seq-KD} = -\mathbbm{E}_{x\sim \mathcal{D}} \sum_{t\in\mathbf{T}} q(t|x) \log p(t|x) \\
    \ = -\mathbbm{E}_{x\sim \mathcal{D},\hat{y}=\mathop{\arg\max}\limits_{t\in\mathbf{T}} q(t|x)} [\log p(t=\hat{y}|x)]
\end{aligned}
\end{equation}

where $t\in\mathcal{T}$ is the space of possible target sequences. This can also be seen as a special case of standard distillation over the sentence space when the temperature $\tau$ approaches 0, which is equivalent to taking the $\arg \max$ over all feasible translations. While the “hard target” $\hat{y}$ is the most likely translation predicted by the teacher, in practice we use beam search as an approximation. As mentioned earlier, almost all the existing literature trains NAT models using sequence-level knowledge distillation from a pre-trained AT model to achieve competitive performance.

\subsection{Self-Distillation Mixup Training}
\label{sec:SDM}

\begin{figure}[htp]
\centering
\includegraphics[width=0.35\textwidth]{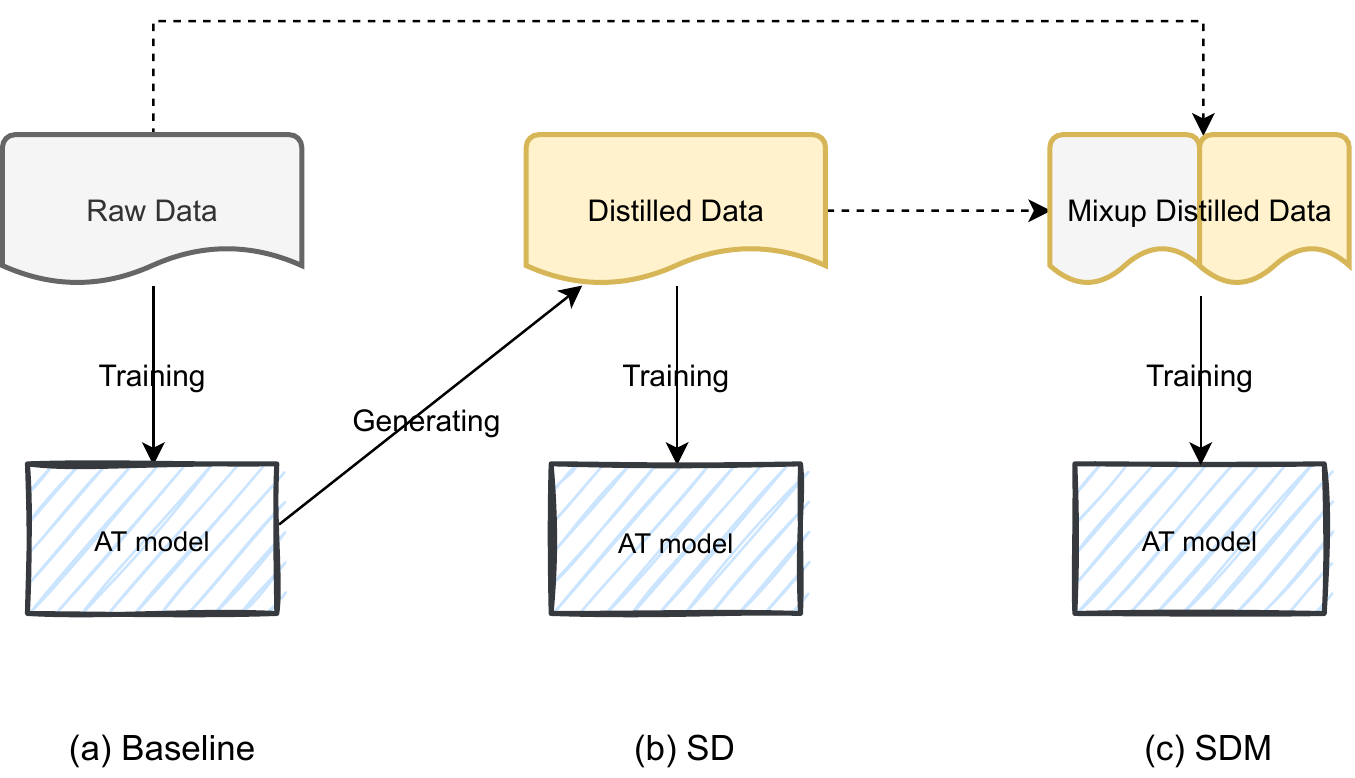} 
\caption{An overview of (a) Baseline Training, (b) Self-Distillation (SD) Training and (c) Self-Distillation Mixup (SDM) Training for an AT model.}
\label{fig:SD_SDM}
\end{figure}

We clarify Self-Distillation Training and Self-Distillation Mixup Training in this section. The overview of these training strategies is shown in Figure \ref{fig:SD_SDM}.

\paragraph{Self-Distillation}
As shown in Figure \ref{fig:SD_SDM} (b), we define Self-Distillation (SD) as training a student model on the distilled data generated by the teacher model which is as same as the student model. 

\paragraph{Self-Distillation Mixup}
The strategy of Mixup Training is transferred from Self-training \citep{DBLP:conf/ijcnn/ArazoOAOM20}. Self-training is an algorithm to train a model to fit pseudo-labels predicted by another previously-learned model. It has been very successful for learning with unlabeled data while mixup training on the combinations of labeled data and pseudo-labeled data.

In NMT task, distilled data can be considered as pseudo-labels. As shown in Figure \ref{fig:SD_SDM} (c), we define Self-Distillation Mixup (SDM) as training the student model on the combinations of original data and distilled data.

Prior research \citet{DBLP:journals/corr/FreitagAS17} has proved that SDM Training can improve the translation quality. We re-implement the results of AT performance on IWSLT14 EN$\rightarrow$DE task, shown in Table \ref{tab:AT_SD_SDM}.

\begin{table}[t]
\centering
\resizebox{0.98\linewidth}{!}{
\begin{tabular}{@{}llcc@{}}
\hline
\textbf{Setup} & \textbf{Paralled Data} & \textbf{BLEU} & $\Delta$\textbf{BLEU} \\
\hline
\hline
Baseline & Raw Data (160K) & {28.34} & - \\
\hline
SD & Distilled Data (160K)  & {28.29} & -0.05 \\
\hline
SDM & Raw Data + & \textbf{29.51} & \textbf{+1.17*}$\uparrow$ \\
 & Distilled Data (320K) & &  \\
\hline
\end{tabular}
}
\caption{Results of AT performance under SD and SDM training strategies on IWSLT14 EN$\rightarrow$DE task.}
\label{tab:AT_SD_SDM}
\end{table}

\section{Problem Formulation}
\subsection{Motivations}
\label{sec:prob_1}
NAT models relay more on the distilled data generated by AT models than the raw data. But prior research \citep{DBLP:conf/iclr/ZhouGN20} has proved that, higher quality distilled data does not necessarily improve the performance of NAT models. V-NAT \citep{DBLP:conf/iclr/Gu0XLS18} achieves better performance on distilled data generated by Transformer-small than Transformer-base/big. CMLM \citep{DBLP:conf/emnlp/GhazvininejadLL19} only achieves a little bit improvement of 0.22 BLEU when trained on distilled data by Transformer-big comparing with Transformer-base.

As mentioned in {\S}\ref{sec:SDM}, a better AT model can be got by using SDM Training. We study several NAT models' performance under AT-baseline teacher model and AT-SDM teacher model. We do this preliminary experiments on IWSLT14 EN$\rightarrow$DE task.

\begin{table}[t]
    \centering
    \resizebox{0.98\linewidth}{!}{
    \begin{tabular}{llcc}
    \hline
    \multicolumn{2}{l}{\textbf{Teacher Model}} & \multicolumn{2}{c}{\textbf{Student Model}} \\
    \textbf{} & \textbf{} & \textbf{NAT-CRF} & \textbf{CMLM} \\
    \hline
    \hline
    AT-Baseline & 28.34 & 22.41 & 26.78 \\
    AT-SDM & 29.51 & 21.83$\downarrow$ & 26.83 $\uparrow$ \\
    \hline
    \end{tabular}
    }
    \caption{Results of several NAT models' performance under AT-baseline teacher model and AT-SDM teacher model on IWSLT14 EN$\rightarrow$DE task.}
    \label{tab:NAT_under_baseline_SDM_teacher}
\end{table}

As shown in Table \ref{tab:NAT_under_baseline_SDM_teacher}, we reach a conclusion consistent with prior research \citep{DBLP:conf/iclr/ZhouGN20} that, higher quality distilled data does not necessarily improve the performance of NAT models. The little bit difference between our experiments and theirs is that we generate distilled data by the same size AT teacher models with the different training strategies, while they use the different size teacher models with the same training strategies. 

\paragraph{Problem 1: } \textbf{SDM can improve the performance of the AT teacher models and lead to higher quality distilled data, but it cannot further improve the performance of NAT student models generally.} We make some theoretical analysis in {\S}\ref{sec:analysis}.

\subsection{Straightforward but Failed Formulation}
\label{sec:prob_2}
\begin{figure}[htp]
\centering
\includegraphics[width=0.40\textwidth]{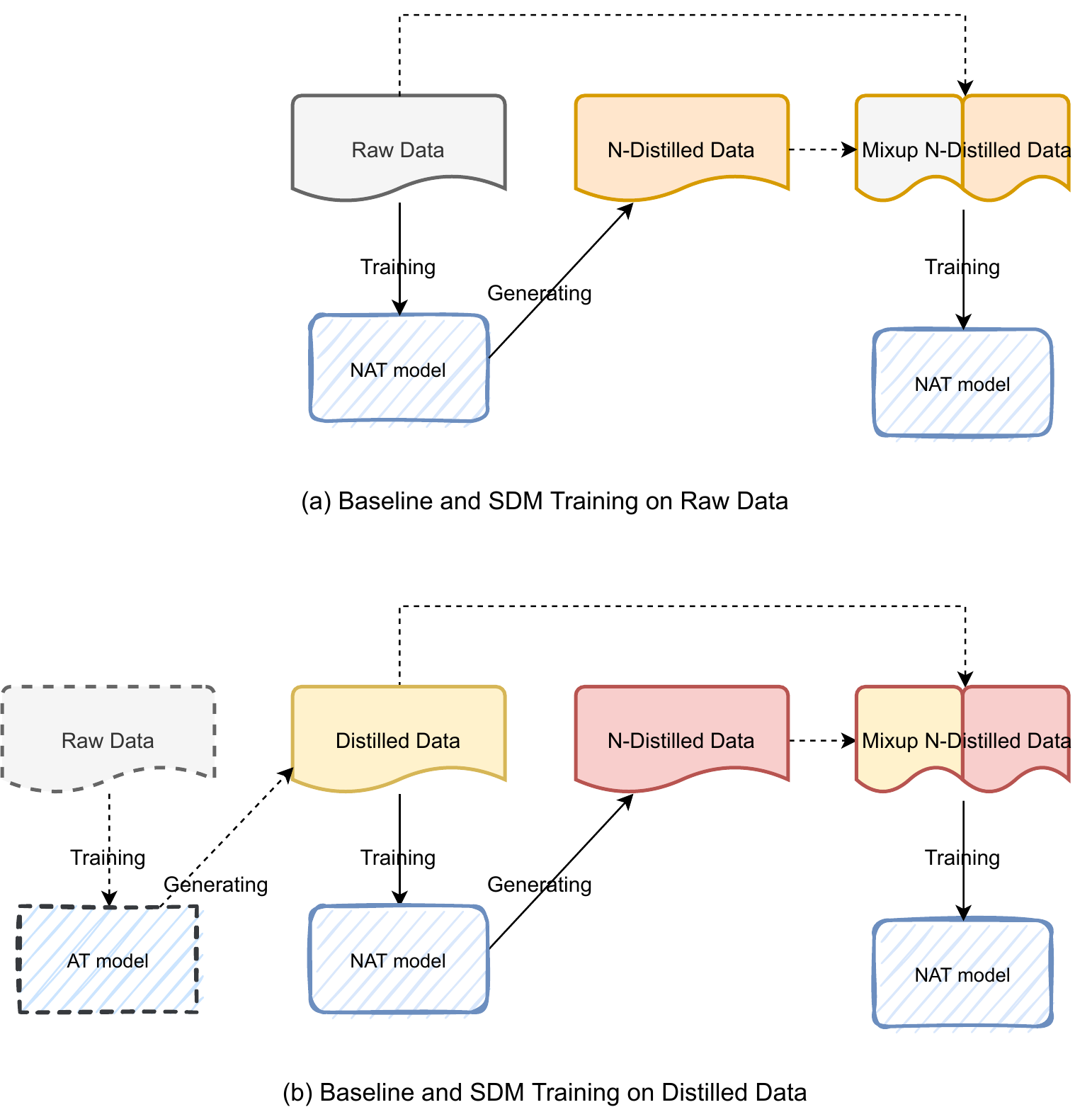} 
\caption{An overview of Baseline Training and SDM Training on (a) Raw Data and (b) Distilled Data for a NAT model.}
\label{fig:NAT_SDM}
\end{figure}

A straightforward way is to directly adopt SDM Training on NAT models. The overview of the training strategies is shown in Figure \ref{fig:NAT_SDM}. We do experiments of all strategies on IWSLT14 EN$\rightarrow$DE task.

\begin{figure*}[htp]
\centering
\includegraphics[width=0.75\textwidth]{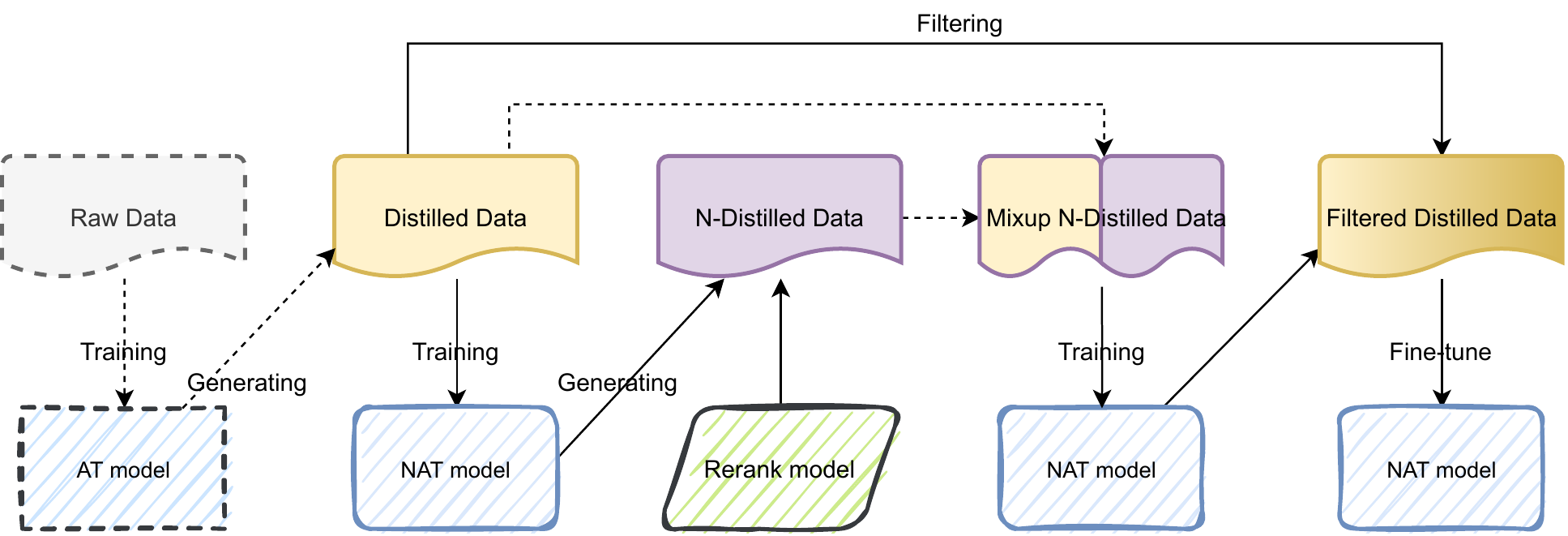} 
\caption{An overview of Self-Distillation Mixup Training with Pre-Rerank and Fine-Tune (SDMRT) Training on distilled data for a NAT model.}
\label{fig:SDMRT}
\end{figure*}


\begin{table}[t]
\centering
\resizebox{0.98\linewidth}{!}{
\begin{tabular}{@{}lcccc@{}}
\hline
\textbf{Arch} & \multicolumn{2}{c}{\textbf{Raw data}} & \multicolumn{2}{c}{\textbf{AT-distilled data}} \\
\textbf{} & \textbf{baseline} & \textbf{SDM} & \textbf{baseline} & \textbf{SDM} \\
\hline
\hline
\textbf{AT} & {28.34} & {29.51} $\uparrow$ & {-} & {-}  \\
\hline
\hline
\textbf{NAT-CRF} & {18.41} & {19.61} $\uparrow$ & {22.41} & {21.93} $\downarrow$\\
\textbf{CMLM} & {25.04} & {25.97} $\uparrow$ & {26.78} & {26.67} $\downarrow$ \\ 
\hline
\end{tabular}
}
\caption{Results of several NAT models’ performance under classic SDM training on IWSLT14 EN$\rightarrow$DE task.}
\label{tab:NAT_with_SDM}
\end{table}

In Table~\ref{tab:NAT_with_SDM}, it is clear that SDM Training can significantly improve the translation quality of NAT models trained on raw data, which has the similar conclusion with AT. But SDM Training gains no improvement when NAT models is trained on distilled data. Such inconsistent results confuse us.

\begin{table}[t]
\centering
\resizebox{0.98\linewidth}{!}{
\begin{tabular}{@{}lcccc@{}}
\hline
\textbf{Arch} & \multicolumn{2}{c}{\textbf{Test Set}} & \multicolumn{2}{c}{\textbf{D-Test Set}} \\
\textbf{} & \textbf{baseline} & \textbf{SDM} & \textbf{baseline} & \textbf{SDM} \\
\hline
\hline
\textbf{NAT-CRF} & {22.41} & {21.93} $\downarrow$ & {36.45} & {36.96} $\uparrow$ \\
\textbf{CMLM} & {26.78} & {26.67} $\downarrow$ & {43.78} & {44.11} $\uparrow$ \\ 
\hline
\end{tabular}
}
\caption{Results of several NAT models’ performance on Test Set and D-Test Set on IWSLT14 EN$\rightarrow$DE task.}
\label{tab:NAT_with_SDM_DTEST}
\end{table}

The only difference is the different \textit{{Train Set}}. One is original, and the other is distilled. But the \textit{{Test Set}} is the same one. So we build a \textit{\textbf{Distilled Test Set}} (\textit{\textbf{D-Test Set}}), which keeps the source side of the test set and replace the target side with the prediction by the AT teacher model. Furthermore, we re-evaluate the performance of the NAT models on D-Test Set and the results in Table \ref{tab:NAT_with_SDM_DTEST} show SDM training worked on D-Test Set. So the key is the diversity between the distilled data and the original data. It is explicable but not expected to the original goal. Given these facts, we conclude that the straightforward formulation is not effective.

\paragraph{Problem 2:} \textbf{How to successfully use SDM Training for NAT models on distilled data, while the distilled data brings bias comparing with the original data?}

\section{Approach}
\label{sec:approach}
\subsection{Self-Distillation Mixup with Pre-Rerank and Fine-Tune Training}
\label{sec:method}

We add two stages to the classic SDM Training process. One is Pre-Rerank on self-distilled data, the other is Fine-Tune on filtered teacher-distilled data. \textbf{The overall training process is seen in Figure} \ref{fig:SDMRT}. We define our enhanced training strategy as Self-Distillation Mixup with Pre-Rerank and Fine-Tune (SDMRT) Training.

Let $\mathcal{D}$ denote the original parallel data: $\mathcal{D} = \{\mathcal{X}; \mathcal{Y}\}$, where $\mathcal{X}$ is the original source-side corpus and $\mathcal{Y}$ is the original target-side corpus. Let $\mathcal{\hat{D}}$ denotes the distilled parallel data: $\mathcal{\hat{D}} = \{\mathcal{X}; \mathcal{\hat{Y}}\}$, where $\mathcal{\hat{Y}}$ is the distilled target-side corpus. For a clearer presentation, we summarise the SDMRT training process concretely in Appendix.

When translation model $M$ generates $k$-candidates $\mathbbm{Y} =\{y_0^{'}, ..., y_{k-1}^{'}\}$ on the input sentence $x$, we can use another model $M_r$ to re-score the candidates instead of scoring them by the model $M$ itself, which is known as Rerank. Reranking is an effective method to change the data distribution. We define the stage of re-score the self-distilled data in our SDMRT as Pre-Rerank. We demonstrate two rerank strategies. One is using Language Model trained on the original target-side corpus $M_r = LM(\mathcal{Y})$ to evaluate the Perplexity of the conditional candidates. The other is using Alignment Model such as an AT teacher model $M_r = AT(\mathcal{X}, \mathcal{Y})$ to score the Conditional Entropy between the source input and the prediction output.

Fine-Tune on filtered teacher-distilled data is an effective method to reduce the bias gap mentioned earlier. The key of the Fine-Tune stage in our SDMRT is to filter incorrect train set out. We calculate the score TER$(\hat{y}, y)$ of the target-side of the teacher-distilled data $\mathcal{\hat{D}}$, where $(x, \hat{y})\in\hat{D}$; $(x, {y})\in{D}$, and only keep the sentence pairs $(x, \hat{y})$ with TER score lower than $\tau$. 

\subsection{Theoretical Analysis}
\label{sec:analysis}
\citep{DBLP:conf/iclr/ZhouGN20} has used Bayesian Decision Theory to analyse distilled data with access to the true distribution. Furthermore, we use Bayesian Decision Theory to explain the reasons for the effectiveness and invalidity of SDM in {\S}\ref{sec:prob_1} and {\S}\ref{sec:prob_2}. We also give the reasons why our SDMRT is effective.

\paragraph{Background: Bayesian Decision Theory}
In the problem of translations, let $x$ denote the input sequence and $y$ denote the output label sequence, where $x \in \mathcal{X} $ and $y \in \mathcal{Y}$. Let $\mathcal{H}$ denote all the possible hypothesis functions from the input to the output space: $\mathcal{H} = \{h: \mathcal{X} \rightarrow \mathcal{Y}\}$. Let $r(y|x)$ denote the conditional risk on the input $x$, which is the expected loss of predicting y based on the posterior probabilities:

\begin{align}
    r(y|x) = \mathbbm{E}_{P(y^\prime|x)}[L(y,y^\prime)] ,
\end{align}

where $L(y,y^\prime)$ is the loss function that penalizes predicting the true target $y^\prime$ as $y$. The classification task aims to find a hypothesis function $h$ that minimizes the overall risk $R$ given by

\begin{align}
    R(h) = \mathbbm{E}_{P(x)}[r(h(x)|x)]
\end{align}

This is known as the Bayes risk. To minimize the overall risk, obviously we need to minimize the conditional risk for each input $x$. The Bayesian decision rule states that the global minimum of $R(h)$ is achieved when the classifier makes predictions that minimize each conditional risk given $x$ and this gives the Bayes optimal classifier:

\begin{align}
    h^*(x) = \mathop{\arg\min}\limits_{y \in \mathcal{Y}} r(y|x)
\end{align}

We define $h^*$ as the best function in theory. So the parallel data can be considered as $\mathcal{D} = \{\mathcal{X}; h^*(\mathcal{X})\}$. So:
\begin{itemize}
 \item For baseline training, we train models on $\{\mathcal{X}; h^*(\mathcal{X})\}$ and get $h_1$ close to $h^*$. 
 \item For SD training, we train models on $\{\mathcal{X}; h_1(\mathcal{X})\}$ and get another $h_1$.
 \item For SDM training, we train models on $\{\mathcal{X}; h^*(\mathcal{X})\} \cup \{\mathcal{X}; h_1(\mathcal{X})\}$ and get a new $h_2$ between $h^*$ and $h_1$, which means $h_2$ is closer to $h^*$.
\end{itemize}

\paragraph{Modeling Diversity}
Let $\mathcal{M}$ denote a trainable network model and $\Theta$ denote the parameters of the model $\mathcal{M}$. Then the $\mathcal{D}$ can be defined as:

\begin{align}
    \mathcal{D} = \{\mathcal{X}; h(\mathcal{X}, M^*, \Theta^*) \}
\end{align}

, where the theoretically best function $h^*$ can be replaced by the theoretically best model $M^*$ and the theoretically best parameters $\Theta^*$. 

Let $M_s$ and $\Theta_{s}$ denote the student model and parameters. Let $M_t$ and $\Theta_{t}$ denote the teacher model and parameters. Then SDM Training can be considered to find the best $(M_{s}, \Theta_{s})$ for:
\begin{align}
    \{\mathcal{X}; h(\mathcal{X}, M^*, \Theta^*)\} \cup \{\mathcal{X}; h(\mathcal{X}, M_{t}^*, \Theta_{t}^*)\}
\end{align}
, where $(M_s, \Theta_{s}) \subseteq (M^*, \Theta^*)$ and $(M_t, \Theta_{t}) \subseteq (M^*, \Theta^*)$. If the model is the same one, the training objective is simplified to only find the best $\Theta_{s}$. So SDM Training is effective when AT or NAT models trained on original data. But when training NAT models on distilled data using SDM, $M_t$ and $M_s$ is different which makes the goal complex.

Modeling Diversity between different models can also be regarded as Domain Specificity \citep{DBLP:conf/naacl/SennrichHB16} between the predictions of these models. We can use Language Model (LM) to evaluate the Perplexity (PPL) of the predictions. We use original target-side of train set to train the LM and evaluate the average Perplexity of test set on IWSLT14 EN$\rightarrow$DE task.

\begin{table}[t]
\centering
\resizebox{0.98\linewidth}{!}{
\begin{tabular}{@{}lcccc@{}}
\hline
\textbf{Metrics} & \textbf{Original} & \multicolumn{3}{c}{\textbf{Model Predictions}} \\
\textbf{} & & \textbf{AT} & \textbf{NAT-CRF} & \textbf{CMLM} \\
\hline
\hline
Perplexity & {23.59} & {28.78} & {127.59} & {37.22}  \\
\hline
\end{tabular}
}
\caption{Average PPL of original Test Set and multiple model predictions on IWSLT14 EN$\rightarrow$DE.}
\label{tab:model_div}
\end{table}

As shown in Table \ref{tab:model_div}, it is clear that different models occur diversity. \textbf{Rerank is an effective method to change the domain distribution of the distilled data generated by $M_s$ to $M_t$.} We further analyse the effect of Rerank in {\S}\ref{sec:analysis_rerank}.

\paragraph{Confirmation Bias}

Network predictions are, of course, sometimes incorrect. In self-training, this situation is reinforced when incorrect predictions are used as labels for unlabeled samples, as it is the case in pseudo-labeling. Overfitting to incorrect pseudo-labels predicted by the network is known as confirmation bias \citep{DBLP:conf/ijcnn/ArazoOAOM20}.

Multi-modality is typically characterized by token repetitions. NAT models may consider many possible translations at the same time and have more likely to generate inconsistent outputs than AT models. We use the frequency of repeated tokens as a proxy for measuring NAT multi-modality \citep{DBLP:conf/icml/GhazvininejadKZ20}. We evaluate the token repetitions of test set under classic SDM training on IWSLT14 EN$\rightarrow$DE task. Multi-modality can be regard as confirmation bias in NMT. 

\begin{table}[t]
\centering
\resizebox{0.98\linewidth}{!}{
\begin{tabular}{@{}lccc@{}}
\hline
\textbf{Metrics} & \multicolumn{3}{c}{\textbf{Model Predictions}} \\
\textbf{} & \textbf{AT} & \textbf{NAT-CRF} & \textbf{CMLM} \\
\hline
\hline
Frequency of  & {0.072\%} & {0.937\%} & {0.461\%}  \\
repeated tokens & - & 13.01$\times$ & 6.42$\times$ \\
\hline
\end{tabular}
}
\caption{The percentage of repeated tokens of different data on IWSLT14 EN$\rightarrow$DE task.}
\label{tab:multi-modality}
\end{table}

\begin{table*}[th]
    \centering
    \resizebox{0.98\linewidth}{!}{
    \begin{tabular}{lcccccc}
    \hline
    \textbf{Model} & \textbf{Iter} & \multicolumn{2}{c}{\textbf{WMT14}} & \multicolumn{2}{c}{\textbf{WMT16}} & \textbf{IWSLT14} \\
     &  & En $\rightarrow$ De & De $\rightarrow$ En & En $\rightarrow$ Ro & Ro $\rightarrow$ En & En $\rightarrow$ De \\
    \hline
    \hline
    \multicolumn{7}{l}{\textbf{AT models}} \\
    \hline
    Transformer-base\citep{DBLP:conf/nips/VaswaniSPUJGKP17} & - & 27.3 & - & - & - & - \\
    Transformer-base(Our Implementation) & - & 27.67 & 31.17 & 34.37 & 33.98 & 28.34 \\
    Transformer-big(Our Implementation) & - & 28.83 & - & - & - & - \\
    \hline
    \hline
    \multicolumn{7}{l}{\textbf{NAT models}} \\
    \hline
    V-NAT\citep{DBLP:conf/iclr/Gu0XLS18} & - & 19.17 & 23.20 & 29.79 & 31.44 & - \\
    NAT-CRF\citep{DBLP:conf/nips/SunLWHLD19} & - & 23.32 & 25.75 & - & - & 26.39 \\
    iNAT\citep{DBLP:conf/emnlp/LeeMC18} & 10 & 21.61 & 25.48 & 29.32 & 30.19 & - \\
    CMLM*\citep{DBLP:conf/emnlp/GhazvininejadLL19} & 10 & 27.03 & 30.53 & 33.08 & 33.31 & - \\
    \hline
    \multicolumn{7}{l}{\textbf{Our Implementation}} \\
    \hline
    V-NAT & - & 16.32 & 20.01 & 24.83 & 25.57 & 15.79 \\
    NAT-CRF & - & 23.76 & 25.56 & 29.31 & 29.17 & 22.41 \\
    iNAT & 10 & 21.19 & 25.37 & 29.17 & 30.09 & 22.75 \\
    CMLM & 10 & 26.03 & 29.61 & 32.94 & 33.07 & 26.78 \\
    CMLM* & 10 & 27.06 & 30.94 & 33.07 & 33.36 & -\\
    \hline
    \multicolumn{7}{l}{\textbf{SDMRT}} \\
    \hline
    V-NAT & - & 17.31 (+0.99) & 21.07 (+1.06) & 25.87 (+1.04) & 26.64 (+1.07) & 16.9 (+1.11) \\
    NAT-CRF & - & 24.68 (+0.92) & 26.47 (+0.91) & 30.35 (+1.04) & 30.26 (+1.09) & 23.51 (+1.10) \\
    iNAT & 10 & 21.97 (+0.78) & 25.94 (+0.57) & 29.94 (+0.77) & 30.9 (+0.81) & 23.64 (+0.89) \\
    CMLM & 10 & \textbf{26.77} (+0.74) & \textbf{30.45} (+0.84) & \textbf{33.57} (+0.63) & \textbf{33.69} (+0.62) & \textbf{27.49} (+0.71) \\
    CMLM* & 10 & \textbf{27.72} (+0.66) & \textbf{31.65} (+0.71)& \textbf{33.72} (+0.65) & \textbf{33.94} (+0.58) & - \\
    \hline
    \end{tabular}
    }
    \caption{Performance of BLEU \citep{DBLP:conf/acl/PapineniRWZ02} score of multiple NAT models on WMT14 En$\leftrightarrow$De, WMT16 En$\leftrightarrow$Ro and IWSLT14 En$\rightarrow$De. Note that we train CMLM baseline using Transformer-base as teacher model, but their \citep{DBLP:conf/emnlp/GhazvininejadLL19} teacher model is Transformer-big. The results of CMLM* are under Transformer-big teacher model.}
    \label{tab:results}
\end{table*}

As shown in Table~\ref{tab:multi-modality}, predictions of NAT models have more repeated tokens than those of AT models, which proves the confirmation bias occurring between the self-distilled data and the teacher-distilled data. \textbf{To reduce the confirmation bias, Fine-Tune the model on the correct data is necessary.} We further analyse the effect of Fine-Tune in {\S}\ref{sec:analysis_filter}.

\section{Experiments \& Results}
We evaluate our SDMRT Training strategy on five standard NMT benchmarks including WMT14 En$\leftrightarrow$De and WMT16 En$\leftrightarrow$Ro in both directions, and IWSLT14 En$\rightarrow$De in single direction.

\subsection{Experiments Setting}
\paragraph{Datasets}
The sizes of the datasets are (train=4.5M / valid=3k / test=3k / dict=42k), (train=610k / valid=2k / test=2k / dict=40k) and (train=160k / valid=7k / test=6k / dict=10k) for WMT14 En$\leftrightarrow$De, WMT14 En$\leftrightarrow$Ro and IWSLT14 En$\rightarrow$De respectively. We use Transformer-base \citep{DBLP:conf/nips/VaswaniSPUJGKP17} as teacher model to create the AT-distilled data, and initialize NAT model with the same model size. We use BPE \citep{DBLP:conf/acl/SennrichHB16a} to tokenize the sentences and create the vocabulary. 

\paragraph{Model Configurations}
For WMT tasks, we use the following hyperparameters: $n_{layers}$ = 12, $n_{heads}$ = 8, $d_{hidden}$ = 512, $d_{FFN}$ = 2048, $lr$ = 5e-4, $n_{warmsup}$ = 10k, $bz$ = 8192, $n_{gpu}$ = 8. For IWSLT tasks, smaller hyperparameters are set: $n_{layers}$ = 12, $n_{heads}$ = 4, $d_{hidden}$ = 512, $d_{FFN}$ = 1024, $lr$ = 5e-4, $n_{warmsup}$ = 6k, $bz$ = 10240, $n_{gpu}$ = 1. Our models are trained on Tesla V100 GPUs. \textit{Adam} \citep{DBLP:journals/corr/KingmaB14} is used as the optimizer and the \textit{inversed-sqrt} scheduler is used for \textit{lr} decaying. All models are trained for 300k steps, and last 4 checkpoints are averaged. For the Pre-Rerank stage, we use AT teacher model to re-score the predictions instead of a language model. For the Fine-Tune stage, we filter the distilled data with TER score less than 0.8, and pre-train models for 240k steps and further-train for 60k steps. We implement models in the experiment with fairseq \citep{ott2019fairseq}.

\subsection{Main Results}
\label{sec:main_results}
Table~\ref{tab:results} shows the translation performance of our SDMRT method. \textbf{For all tasks, SDMRT forms the baselines by a large margin (0.6$\sim$1.2 BLEU) on distilled data.}

\subsection{Study}
\paragraph{SDM v.s. SDMRT on different NAT models} As shown in Table~\ref{tab:SDMvsSDMRT}, the performance of our SDMRT outperforms that of classic SDM. Surprisingly, NAT-CRF with SDM can achieve positive results on smaller datasets, while CMLM achieves positive results on larger datasets.

\begin{table}[t]
\centering
\resizebox{0.98\linewidth}{!}{
\begin{tabular}{@{}lcc@{}}
\hline
\textbf{Model} & \textbf{WMT14 EN$\rightarrow$DE} & \textbf{IWSLT14 EN$\rightarrow$DE} \\
\hline
\hline
\multicolumn{3}{l}{\textbf{Baseline}} \\
\hline
NAT-CRF & {23.76} & {22.41} \\
CMLM & {26.03} & {26.78} \\
\hline
\multicolumn{3}{l}{\textbf{SDM}} \\
\hline
NAT-CRF & {23.15(-0.61)} & {22.95(+0.54)} \\
CMLM & {26.23(+0.20)} & {26.67(-0.11)} \\
\hline
\multicolumn{3}{l}{\textbf{SDMRT}} \\
\hline
NAT-CRF & {24.68(+0.92)} & {23.51(+1.10)} \\
CMLM & {26.77(+0.74)} & {27.49(+0.71)} \\
\hline
\end{tabular}
}
\caption{SDM v.s. SDMRT on IWSLT14 EN$\rightarrow$DE task and WMT14 EN$\rightarrow$DE task.}
\label{tab:SDMvsSDMRT}
\end{table}

\begin{table}[t]
\centering
\resizebox{0.80\linewidth}{!}{
\begin{tabular}{@{}lccc@{}}
\hline
\textbf{STEP} & \textbf{Baseline} & \textbf{SDMRT} & \textbf{$\Delta$BLEU} \\
\hline
\hline
1 & {19.54} & {21.58} & {2.04} \\
5 & {25.57} & \textbf{26.24*} & {0.67} \\
10 & \textbf{26.03} & \textbf{26.77*} & {0.74} \\
\hline
\end{tabular}
}
\caption{BLEU scores of every step with maxiter=10 for CMLM model on WMT14 EN$\rightarrow$DE task. SDMRT can overstep baseline earlier in step of 5.}
\label{tab:acceletation}
\end{table}

\begin{figure}[htp]
\centering
\includegraphics[width=0.50\textwidth]{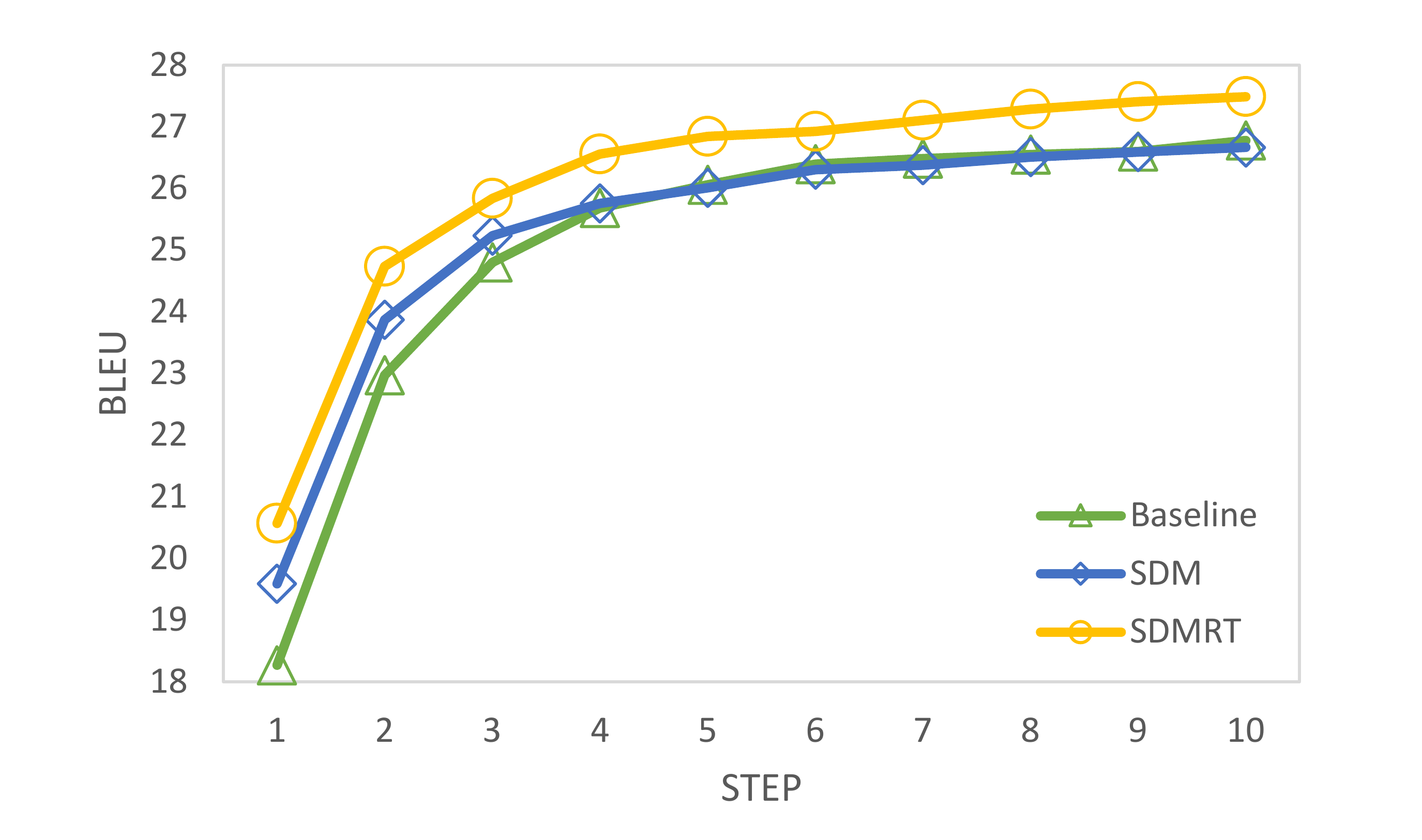} 
\caption{Results of different training strategies for CMLM model on IWSLT14 EN$\rightarrow$DE task.}
\label{fig:CMLM_SDM_SDMRT}
\end{figure}

\paragraph{SDM v.s. SDMRT on different steps of IR NAT models} It is clear in Figure~\ref{fig:CMLM_SDM_SDMRT}, SDM only has positive influence at the first several steps, but our SDMRT can improve translation quality at any step.

\paragraph{Acceleration for Iterative Refinement NAT} 
\label{sec:speed_irnat}
For Iterative Refinement  (IR) NAT such as CMLM, it usually requires more than 4 steps to make the refinement converge, which causes negative influence on the speedup. Table~\ref{tab:acceletation} shows, on AT-distilled data, our SDMRT training strategy can overstep baseline within 5 steps which means 2$\times$ acceleration compared to baseline.

\begin{table}[t]
\centering
\resizebox{1\linewidth}{!}{
\begin{tabular}{@{}lcccc@{}}
\hline
\textbf{Metrics} & \textbf{Rerank} & \textbf{Original} & \textbf{NAT-CRF} & \textbf{CMLM} \\
\hline
\hline
Perplexity & {w/o} & {23.59}  & {127.59} & {37.22}  \\
 & {w/} & {-} & {76.55}$\downarrow$ & {30.47}$\downarrow$ \\
\hline
\end{tabular}
}
\caption{Re-visit average PPL of multiple models predictions on IWSLT14 EN$\rightarrow$DE.}
\label{tab:model_div-revisit}
\end{table}

\begin{table}[t]
\centering
\resizebox{0.98\linewidth}{!}{
\begin{tabular}{@{}lcccc@{}}
\hline
\textbf{Model} & \multicolumn{2}{c}{\textbf{SDM}} & \multicolumn{2}{c}{\textbf{SDMRT}} \\
\hline
\hline
AT & {0.072\%} & - & - & - \\
\hline
NAT-CRF & {0.937\%} & 13.01$\times$ & {0.533\%}$\downarrow$ & 7.40$\times$ \\
CMLM & {0.461\%} & 6.42$\times$ & {0.195\%}$\downarrow$ & 2.71$\times$ \\
\hline
\end{tabular}
}
\caption{Re-visit the percentage of repeated tokens of different data on IWSLT14 EN$\rightarrow$DE task.}
\label{tab:multi-modality-revisit}
\end{table}

\begin{table}[t]
\centering
\resizebox{0.80\linewidth}{!}{
\begin{tabular}{@{}lcc@{}}
\hline
\textbf{Model}  & \multicolumn{2}{c}{\textbf{SDMRT}} \\
 & \textbf{w/ LM-Rerank} & \textbf{w/ AT-Rerank} \\
\hline
\hline
NAT-CRF  & \textbf{24.72} & {24.68}\\
CMLM  & {26.33} & \textbf{26.77} \\
\hline
\end{tabular}
}
\caption{Different Rerank strategies on WMT14 EN$\rightarrow$DE task.}
\label{tab:ATvsLM}
\end{table}

\begin{table}[t]
\centering
\resizebox{0.98\linewidth}{!}{
\begin{tabular}{@{}lcc@{}}
\hline
\textbf{Model} & \multicolumn{2}{c}{\textbf{SDMRT}} \\
 & \textbf{Data-Filtering} & \textbf{No Data-Filtering} \\
\hline
\hline
NAT-CRF & {24.68} & {23.83} (-0.85)\\
CMLM & {26.77} & {26.13} (-0.64)\\
\hline
\end{tabular}
}
\caption{Different Data-Filtering strategies on WMT14 EN$\rightarrow$DE task.}
\label{tab:data-filter}
\end{table}

\paragraph{Re-visit Modeling Specificity}
\label{sec:revisit-modeling}

Following the preliminary experiments before, we re-visit the average Perplexity of test set on IWSLT14 EN$\rightarrow$DE. The results in Table~\ref{tab:model_div-revisit} prove that the distribution of NAT models' predictions under Rerank is more closer to that of original data.

\paragraph{Re-visit Confirmation Bias}
\label{sec:revisit-bias}
As mentioned earlier, confirmation bias in NMT is Multi-modality, and Multi-modality is typically characterized by token repetitions. We re-measure the frequency of repeated tokens. As shown in Table~\ref{tab:multi-modality-revisit}, the percentage of repeated tokens of with rerank stage sharply decreases, which proves that our SDMRT is an effective method.

\paragraph{AT-Rerank v.s. LM-Rerank}
\label{sec:analysis_rerank}
We compare the effect of different rerank strategies. As shown in Table~\ref{tab:ATvsLM}, AT-Rerank performs overall better than LM-Rerank. Specially, LM-Rerank can achieve similar performance as AT-Rerank for NAT-CRF model.

\paragraph{Effect of Data-Filtering on Fine-Tune Stage}
\label{sec:analysis_filter}
To measure the effect of filtered teacher-distilled data, we change our SDMRT training process and fine-tune the models on teacher-distilled data. The results in Table~\ref{tab:data-filter} prove that Data-Filtering is important in our SDMRT training process.

\section{Conclusion}

In this work, we empirically find directly adopting the classic Self-Distillation Mixup (SDM) Training strategy to NAT models is not as effective as in AT models.
Through careful analysis, we observe $\textit{Modeling Diversity}$ and $\textit{Confirmation Bias}$ between the AT teacher model and the NAT student models, 
causing SDM theoretically invalid for NAT models when trained on AT-distilled data.
Furthermore, we propose an enhanced training strategy named SDMRT and achieve new SOTA performance for multiple NAT models compared to the baselines. 
In addition, it's demonstrated to be an effective method to reduce the iteration number for Iterative Refinement NAT.

\bibliography{acl2020,custom}

\begin{thebibliography}{21}
\expandafter\ifx\csname natexlab\endcsname\relax\def\natexlab#1{#1}\fi

\bibitem[{Arazo et~al.(2020)Arazo, Ortego, Albert, O'Connor, and
  McGuinness}]{DBLP:conf/ijcnn/ArazoOAOM20}
Eric Arazo, Diego Ortego, Paul Albert, Noel~E. O'Connor, and Kevin McGuinness.
  2020.
\newblock \href {https://doi.org/10.1109/IJCNN48605.2020.9207304}
  {Pseudo-labeling and confirmation bias in deep semi-supervised learning}.
\newblock In \emph{2020 International Joint Conference on Neural Networks,
  {IJCNN} 2020, Glasgow, United Kingdom, July 19-24, 2020}, pages 1--8. {IEEE}.

\bibitem[{Bahdanau et~al.(2015)Bahdanau, Cho, and
  Bengio}]{DBLP:journals/corr/BahdanauCB14}
Dzmitry Bahdanau, Kyunghyun Cho, and Yoshua Bengio. 2015.
\newblock \href {http://arxiv.org/abs/1409.0473} {Neural machine translation by
  jointly learning to align and translate}.
\newblock In \emph{3rd International Conference on Learning Representations,
  {ICLR} 2015, San Diego, CA, USA, May 7-9, 2015, Conference Track
  Proceedings}.

\bibitem[{Du et~al.(2021)Du, Tu, and Jiang}]{DBLP:conf/icml/DuTJ21}
Cunxiao Du, Zhaopeng Tu, and Jing Jiang. 2021.
\newblock \href {http://proceedings.mlr.press/v139/du21c.html} {Order-agnostic
  cross entropy for non-autoregressive machine translation}.
\newblock In \emph{Proceedings of the 38th International Conference on Machine
  Learning, {ICML} 2021, 18-24 July 2021, Virtual Event}, volume 139 of
  \emph{Proceedings of Machine Learning Research}, pages 2849--2859. {PMLR}.

\bibitem[{Freitag et~al.(2017)Freitag, Al{-}Onaizan, and
  Sankaran}]{DBLP:journals/corr/FreitagAS17}
Markus Freitag, Yaser Al{-}Onaizan, and Baskaran Sankaran. 2017.
\newblock \href {http://arxiv.org/abs/1702.01802} {Ensemble distillation for
  neural machine translation}.
\newblock \emph{CoRR}, abs/1702.01802.

\bibitem[{Ghazvininejad et~al.(2020)Ghazvininejad, Karpukhin, Zettlemoyer, and
  Levy}]{DBLP:conf/icml/GhazvininejadKZ20}
Marjan Ghazvininejad, Vladimir Karpukhin, Luke Zettlemoyer, and Omer Levy.
  2020.
\newblock \href {http://proceedings.mlr.press/v119/ghazvininejad20a.html}
  {Aligned cross entropy for non-autoregressive machine translation}.
\newblock In \emph{Proceedings of the 37th International Conference on Machine
  Learning, {ICML} 2020, 13-18 July 2020, Virtual Event}, volume 119 of
  \emph{Proceedings of Machine Learning Research}, pages 3515--3523. {PMLR}.

\bibitem[{Ghazvininejad et~al.(2019)Ghazvininejad, Levy, Liu, and
  Zettlemoyer}]{DBLP:conf/emnlp/GhazvininejadLL19}
Marjan Ghazvininejad, Omer Levy, Yinhan Liu, and Luke Zettlemoyer. 2019.
\newblock \href {https://doi.org/10.18653/v1/D19-1633} {Mask-predict: Parallel
  decoding of conditional masked language models}.
\newblock In \emph{Proceedings of the 2019 Conference on Empirical Methods in
  Natural Language Processing and the 9th International Joint Conference on
  Natural Language Processing, {EMNLP-IJCNLP} 2019, Hong Kong, China, November
  3-7, 2019}, pages 6111--6120. Association for Computational Linguistics.

\bibitem[{Gu et~al.(2018)Gu, Bradbury, Xiong, Li, and
  Socher}]{DBLP:conf/iclr/Gu0XLS18}
Jiatao Gu, James Bradbury, Caiming Xiong, Victor O.~K. Li, and Richard Socher.
  2018.
\newblock \href {https://openreview.net/forum?id=B1l8BtlCb} {Non-autoregressive
  neural machine translation}.
\newblock In \emph{6th International Conference on Learning Representations,
  {ICLR} 2018, Vancouver, BC, Canada, April 30 - May 3, 2018, Conference Track
  Proceedings}. OpenReview.net.

\bibitem[{Gu and Kong(2021)}]{DBLP:conf/acl/GuK21}
Jiatao Gu and Xiang Kong. 2021.
\newblock \href {https://doi.org/10.18653/v1/2021.findings-acl.11} {Fully
  non-autoregressive neural machine translation: Tricks of the trade}.
\newblock In \emph{Findings of the Association for Computational Linguistics:
  {ACL/IJCNLP} 2021, Online Event, August 1-6, 2021}, volume {ACL/IJCNLP} 2021
  of \emph{Findings of {ACL}}, pages 120--133. Association for Computational
  Linguistics.

\bibitem[{Hinton et~al.(2015)Hinton, Vinyals, and
  Dean}]{DBLP:journals/corr/HintonVD15}
Geoffrey~E. Hinton, Oriol Vinyals, and Jeffrey Dean. 2015.
\newblock \href {http://arxiv.org/abs/1503.02531} {Distilling the knowledge in
  a neural network}.
\newblock \emph{CoRR}, abs/1503.02531.

\bibitem[{Kim and Rush(2016)}]{DBLP:conf/emnlp/KimR16}
Yoon Kim and Alexander~M. Rush. 2016.
\newblock \href {https://doi.org/10.18653/v1/d16-1139} {Sequence-level
  knowledge distillation}.
\newblock In \emph{Proceedings of the 2016 Conference on Empirical Methods in
  Natural Language Processing, {EMNLP} 2016, Austin, Texas, USA, November 1-4,
  2016}, pages 1317--1327. The Association for Computational Linguistics.

\bibitem[{Kingma and Ba(2015)}]{DBLP:journals/corr/KingmaB14}
Diederik~P. Kingma and Jimmy Ba. 2015.
\newblock \href {http://arxiv.org/abs/1412.6980} {Adam: {A} method for
  stochastic optimization}.
\newblock In \emph{3rd International Conference on Learning Representations,
  {ICLR} 2015, San Diego, CA, USA, May 7-9, 2015, Conference Track
  Proceedings}.

\bibitem[{Lee et~al.(2021)Lee, Auli, and Ranzato}]{DBLP:conf/acl/0001AR20}
Ann Lee, Michael Auli, and Marc'Aurelio Ranzato. 2021.
\newblock \href {https://doi.org/10.18653/v1/2021.acl-long.563} {Discriminative
  reranking for neural machine translation}.
\newblock In \emph{Proceedings of the 59th Annual Meeting of the Association
  for Computational Linguistics and the 11th International Joint Conference on
  Natural Language Processing, {ACL/IJCNLP} 2021, (Volume 1: Long Papers),
  Virtual Event, August 1-6, 2021}, pages 7250--7264. Association for
  Computational Linguistics.

\bibitem[{Lee et~al.(2018)Lee, Mansimov, and Cho}]{DBLP:conf/emnlp/LeeMC18}
Jason Lee, Elman Mansimov, and Kyunghyun Cho. 2018.
\newblock \href {https://doi.org/10.18653/v1/d18-1149} {Deterministic
  non-autoregressive neural sequence modeling by iterative refinement}.
\newblock In \emph{Proceedings of the 2018 Conference on Empirical Methods in
  Natural Language Processing, Brussels, Belgium, October 31 - November 4,
  2018}, pages 1173--1182. Association for Computational Linguistics.

\bibitem[{Ott et~al.(2019)Ott, Edunov, Baevski, Fan, Gross, Ng, Grangier, and
  Auli}]{ott2019fairseq}
Myle Ott, Sergey Edunov, Alexei Baevski, Angela Fan, Sam Gross, Nathan Ng,
  David Grangier, and Michael Auli. 2019.
\newblock fairseq: A fast, extensible toolkit for sequence modeling.
\newblock In \emph{Proceedings of NAACL-HLT 2019: Demonstrations}.

\bibitem[{Papineni et~al.(2002)Papineni, Roukos, Ward, and
  Zhu}]{DBLP:conf/acl/PapineniRWZ02}
Kishore Papineni, Salim Roukos, Todd Ward, and Wei{-}Jing Zhu. 2002.
\newblock \href {https://doi.org/10.3115/1073083.1073135} {Bleu: a method for
  automatic evaluation of machine translation}.
\newblock In \emph{Proceedings of the 40th Annual Meeting of the Association
  for Computational Linguistics, July 6-12, 2002, Philadelphia, PA, {USA}},
  pages 311--318. {ACL}.

\bibitem[{Saharia et~al.(2020)Saharia, Chan, Saxena, and
  Norouzi}]{DBLP:conf/emnlp/SahariaCSN20}
Chitwan Saharia, William Chan, Saurabh Saxena, and Mohammad Norouzi. 2020.
\newblock \href {https://doi.org/10.18653/v1/2020.emnlp-main.83}
  {Non-autoregressive machine translation with latent alignments}.
\newblock In \emph{Proceedings of the 2020 Conference on Empirical Methods in
  Natural Language Processing, {EMNLP} 2020, Online, November 16-20, 2020},
  pages 1098--1108. Association for Computational Linguistics.

\bibitem[{Sennrich et~al.(2016{\natexlab{a}})Sennrich, Haddow, and
  Birch}]{DBLP:conf/naacl/SennrichHB16}
Rico Sennrich, Barry Haddow, and Alexandra Birch. 2016{\natexlab{a}}.
\newblock \href {https://doi.org/10.18653/v1/n16-1005} {Controlling politeness
  in neural machine translation via side constraints}.
\newblock In \emph{{NAACL} {HLT} 2016, The 2016 Conference of the North
  American Chapter of the Association for Computational Linguistics: Human
  Language Technologies, San Diego California, USA, June 12-17, 2016}, pages
  35--40. The Association for Computational Linguistics.

\bibitem[{Sennrich et~al.(2016{\natexlab{b}})Sennrich, Haddow, and
  Birch}]{DBLP:conf/acl/SennrichHB16a}
Rico Sennrich, Barry Haddow, and Alexandra Birch. 2016{\natexlab{b}}.
\newblock \href {https://doi.org/10.18653/v1/p16-1162} {Neural machine
  translation of rare words with subword units}.
\newblock In \emph{Proceedings of the 54th Annual Meeting of the Association
  for Computational Linguistics, {ACL} 2016, August 7-12, 2016, Berlin,
  Germany, Volume 1: Long Papers}. The Association for Computer Linguistics.

\bibitem[{Sun et~al.(2019)Sun, Li, Wang, He, Lin, and
  Deng}]{DBLP:conf/nips/SunLWHLD19}
Zhiqing Sun, Zhuohan Li, Haoqing Wang, Di~He, Zi~Lin, and Zhi{-}Hong Deng.
  2019.
\newblock \href
  {https://proceedings.neurips.cc/paper/2019/hash/74563ba21a90da13dacf2a73e3ddefa7-Abstract.html}
  {Fast structured decoding for sequence models}.
\newblock In \emph{Advances in Neural Information Processing Systems 32: Annual
  Conference on Neural Information Processing Systems 2019, NeurIPS 2019,
  December 8-14, 2019, Vancouver, BC, Canada}, pages 3011--3020.

\bibitem[{Vaswani et~al.(2017)Vaswani, Shazeer, Parmar, Uszkoreit, Jones,
  Gomez, Kaiser, and Polosukhin}]{DBLP:conf/nips/VaswaniSPUJGKP17}
Ashish Vaswani, Noam Shazeer, Niki Parmar, Jakob Uszkoreit, Llion Jones,
  Aidan~N. Gomez, Lukasz Kaiser, and Illia Polosukhin. 2017.
\newblock \href
  {https://proceedings.neurips.cc/paper/2017/hash/3f5ee243547dee91fbd053c1c4a845aa-Abstract.html}
  {Attention is all you need}.
\newblock In \emph{Advances in Neural Information Processing Systems 30: Annual
  Conference on Neural Information Processing Systems 2017, December 4-9, 2017,
  Long Beach, CA, {USA}}, pages 5998--6008.

\bibitem[{Zhou et~al.(2020)Zhou, Gu, and Neubig}]{DBLP:conf/iclr/ZhouGN20}
Chunting Zhou, Jiatao Gu, and Graham Neubig. 2020.
\newblock \href {https://openreview.net/forum?id=BygFVAEKDH} {Understanding
  knowledge distillation in non-autoregressive machine translation}.
\newblock In \emph{8th International Conference on Learning Representations,
  {ICLR} 2020, Addis Ababa, Ethiopia, April 26-30, 2020}. OpenReview.net.

\end{thebibliography}
\bibliographystyle{acl_natbib}

\appendix
\section{Appendix}
\begin{algorithm*}[htp]
\begin{algorithmic}
\STATE 1: \textbf{procedure} TRAIN\_RERANK($\mathcal{D} = \{\mathcal{X}; \mathcal{Y}\}$): \\
\STATE 2: \hspace{\algorithmicindent} \textbf{If} ( $M_r$ is Language Model ) \\
\STATE 3: \hspace{\algorithmicindent} \hspace{\algorithmicindent} Train randomly initialized $M_r$ on $\mathcal{Y}$  \\
\STATE 4: \hspace{\algorithmicindent} \textbf{Elseif} ( $M_r$ is Alignment Model ) \\
\STATE 5: \hspace{\algorithmicindent} \hspace{\algorithmicindent} Train randomly initialized $M_r$ on $\mathcal{D}$  \\
\STATE 6: \hspace{\algorithmicindent} \textbf{Return} $M_r$ \\
\STATE 7: \\
\STATE 1: \textbf{procedure} FILTER($\mathcal{D} = \{\mathcal{X}; \mathcal{Y}\}$, $\mathcal{\hat{D}} = \{\mathcal{X}; \mathcal{\hat{Y}}\}$): \\
\STATE 2: \hspace{\algorithmicindent}  Filter $\mathcal{D}^*=$ \{($x$, $\hat{y}$) $|$ TER($\hat{y}$, $y$) $<\tau$; $(x, \hat{y})\in\hat{D}$; $(x, {y})\in{D}$ \}  \\
\STATE 3: \hspace{\algorithmicindent} \textbf{Return} $\mathcal{D}^*$ \\
\STATE 4: \\
\STATE 1: \textbf{procedure} TRAIN($\mathcal{D} = \{\mathcal{X}; \mathcal{Y}\}$): \\
\STATE 2: \hspace{\algorithmicindent} Train randomly initialized $M$ on $\mathcal{D}$  \\
\STATE 3: \hspace{\algorithmicindent} \textbf{Return} $M$ \\
\STATE 4: \\
\STATE 1: \textbf{procedure} RERANK\_DIS($\mathcal{D} = \{\mathcal{X}; \mathcal{Y}\}$, $M$, $M_r$): \\
\STATE 2: \hspace{\algorithmicindent} \textbf{For} ($x$, $y$) in $\mathcal{D}$:  \\
\STATE 3: \hspace{\algorithmicindent} \hspace{\algorithmicindent} Generate $k$-candidates $\mathbbm{Y} =\{y_0^{'}, ..., y_{k-1}^{'}\}$ by $M(x)$  \\
\STATE 4: \hspace{\algorithmicindent} \hspace{\algorithmicindent} \textbf{If} ( $M_r$ is Language Model ) \\
\STATE 5: \hspace{\algorithmicindent} \hspace{\algorithmicindent} \hspace{\algorithmicindent} Select the best $y^{'}$ order by sorted $\{M_r(y^{'})|y^{'}\in\mathbbm{Y}\}$ \\
\STATE 6: \hspace{\algorithmicindent} \hspace{\algorithmicindent} \textbf{Elseif} ( $M_r$ is Alignment Model ) \\
\STATE 7: \hspace{\algorithmicindent} \hspace{\algorithmicindent} \hspace{\algorithmicindent} Select the best $y^{'}$ order by sorted $\{M_r(x, y^{'})|y^{'}\in\mathbbm{Y}\}$ \\
\STATE 8: \hspace{\algorithmicindent} \textbf{Return} $\mathcal{D}^{'} = \{(x, y^{'})\}$ \\
\STATE 9: \\
\STATE 1: \textbf{procedure} FINE\_TUNE($M$, $\mathcal{D} = \{\mathcal{X}; \mathcal{Y}\}$): \\
\STATE 2: \hspace{\algorithmicindent} Further train $M$ on $\mathcal{D}$  \\
\STATE 3: \hspace{\algorithmicindent} \textbf{Return} $M$ \\
\STATE 4: \\
\STATE 1: \textbf{procedure} SDMRT($\mathcal{D} = \{\mathcal{X}; \mathcal{Y}\}$, $\mathcal{\hat{D}} = \{\mathcal{X}; \mathcal{\hat{Y}}\}$): \\
\STATE 2: \hspace{\algorithmicindent} $M_{r}\leftarrow$TRAIN\_RERANK($\mathcal{D}$) \\
\STATE 3: \hspace{\algorithmicindent} $\mathcal{D}^*\leftarrow$ FILTER($\mathcal{D}$, $\mathcal{\hat{D}}$) 
\STATE 4: \hspace{\algorithmicindent} $M\leftarrow$TRAIN($\mathcal{\hat{D}}$)  \\
\STATE 5: \hspace{\algorithmicindent} $\hat{\mathcal{D}}_{N}\leftarrow$RERANK\_DIS($\mathcal{D}$, $M$, $M_r$) \\
\STATE 6: \hspace{\algorithmicindent} $M\leftarrow$TRAIN($\mathcal{\hat{D}}\cup\hat{\mathcal{D}_{N}}$)  \\
\STATE 7: \hspace{\algorithmicindent} $M\leftarrow$FINE\_TUNE($M, \hat{\mathcal{D}^{*}}$)  \\
\STATE 8: \hspace{\algorithmicindent} \textbf{Return} $M$ \\
\end{algorithmic}
\caption{SDMRT Training}
\label{algo:SDMRT}
\end{algorithm*}

\end{document}